\documentclass[letterpaper]{article}
\usepackage{aaai}
\usepackage{times}
\usepackage{epsfig}
\usepackage{graphicx}
\usepackage{amsmath}
\usepackage{amssymb}
\usepackage{amsfonts}
\usepackage{amsmath,mathtools}
\usepackage{algpseudocode}
\usepackage{color}
\usepackage{multirow}
\usepackage{subfig}
\usepackage{graphicx}

\usepackage[ruled, vlined, linesnumbered]{algorithm2e}

\newcommand{\M}[1]{$\mathcal{M}$}

\begin{document}
%
\title{Interactive Surveillance Technologies for Dense Crowds}
\author{Aniket Bera\\
Department of Computer Science\\
University of North Carolina at Chapel Hill\\
Chapel Hill, NC\\
\And
Dinesh Manocha\\
Department of Computer Science\\
University of Maryland at College Park\\
College Park, MD\\
}

\maketitle
\begin{abstract}
\begin{quote}
We present an algorithm for realtime anomaly detection in low to medium density crowd videos using trajectory-level behavior learning. Our formulation combines online tracking algorithms from computer vision, non-linear pedestrian motion models from crowd simulation, and Bayesian learning techniques to automatically compute the trajectory-level pedestrian behaviors for each agent in the video. These learned behaviors are used to segment the trajectories and motions of different pedestrians or agents and detect anomalies.  We demonstrate the interactive performance on the PETS ARENA dataset as well as indoor and outdoor crowd video benchmarks consisting of tens of human agents. We also discuss the implications of recent public policy and law enforcement issues relating to surveillance and our research.
\end{quote}
\end{abstract}

\section{Introduction}

\begin{figure}[ht]
  \centering
  \includegraphics[width = 1.0\linewidth]{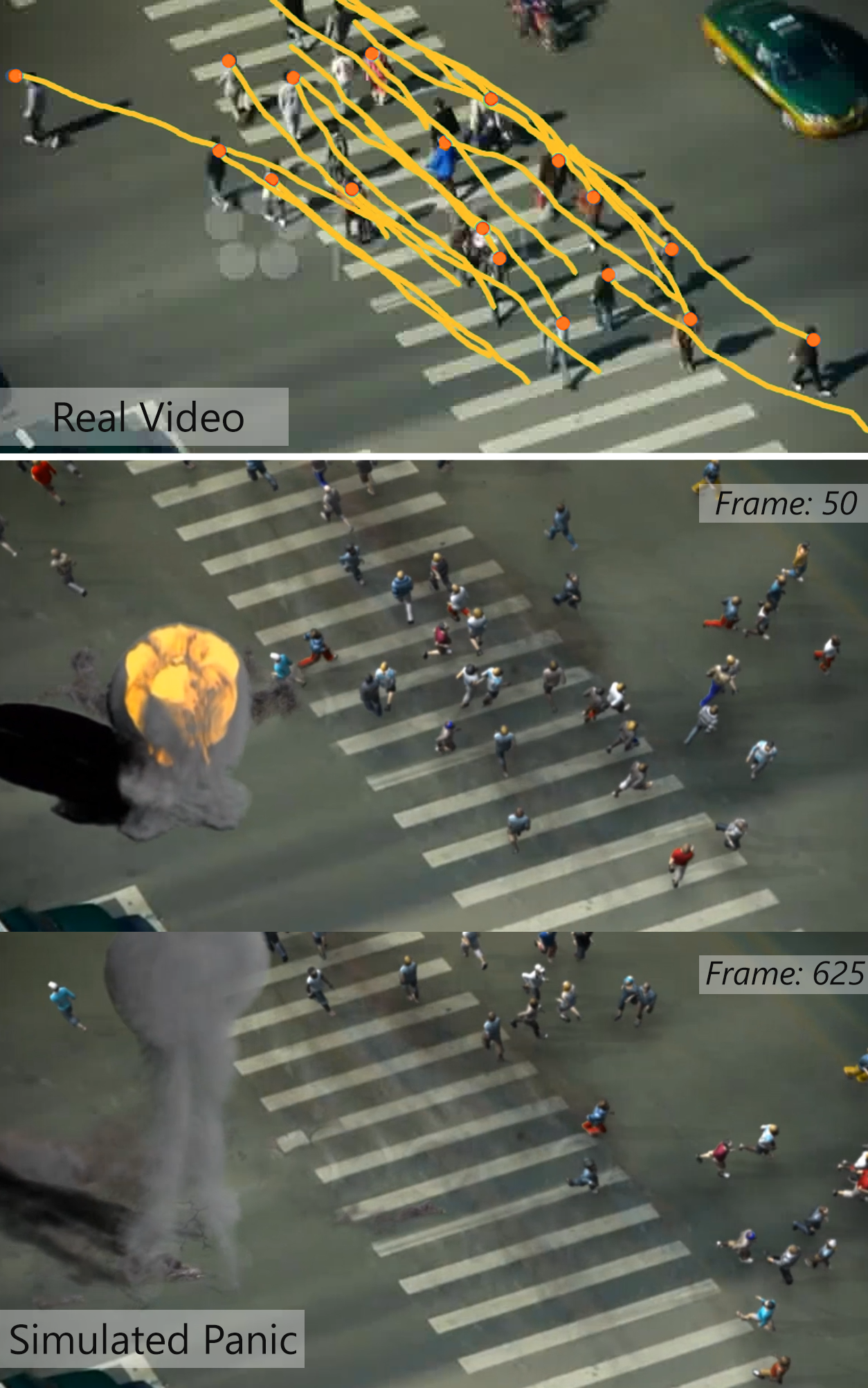}
\caption{ Our method extracts trajectories and computes pedestrian movement features at interactive rates. We use the learned behavior and movement features to detect anomalies in the pedestrian trajectories. The lines indicate behavior features (explained in detail in Section 3.5). The yellow lines indicate anomalies detected by our approach.} \label{fig:cover}
\end{figure}

There has been a growing interest in developing computational methodologies for simulating and analyzing the movements and behaviors of crowds in real-world videos. This include simulation of large crowds composed of a large number of pedestrians or agents, moving in a shared space, and interacting with each other. Some of the driving applications include surveillance, training systems, robotics, navigation, computer games, and urban planning. 

In this paper, we deal with the problem of interactive anomaly detection in crowd videos and develop approaches that perform no precomputation or offline learning. Our research is motivated by the widespread use of commodity cameras that are increasingly used for surveillance and monitoring, including sporting events, public places, religious and political gatherings, etc. One of the key challenges is to devise methods that can automatically analyze the behavior and movement patterns in crowd videos to detect anomalous or atypical behaviors~\cite{LiCrowdedSceneAnalysis2015}. Furthermore, many of these applications desire interactive or realtime performance, and do not rely on apriori learning or labeling. Many algorithms have been designed to track individual agents and/or to recognize their behavior and movements and detect abnormal behaviors)~\cite{Junior2010}. However, current methods are typically limited to sparse crowds or are designed for offline or non-realtime applications.

We present an algorithm for realtime anomaly detection in low to medium density crowd videos. Our approach uses online methods to track each pedestrian and learn the trajectory-level behaviors for each agent by combining non-linear motion models and Bayesian learning. Given a video stream, we extract the trajectory of each agent using a realtime multi-person tracking algorithm that can model different interactions between the pedestrians and the obstacles. Next, we use a Bayesian inference technique to compute the trajectory behavior feature for each agent. These  trajectory behavior features are used for anomaly detection in terms of pedestrian movement or behaviors.  Our approach involves no offline learning and can be used for interactive surveillance and any crowd videos. 
We refer the readers to read~\cite{Bera_2016_CVPR_Workshops} for more technical details and analysis. 







\section{Related Work}

There is extensive research in computer vision and multimedia analyzing crowd behaviors and movements from videos~\cite{LiCrowdedSceneAnalysis2015}. Most of the work has focused on extracting useful information including  behavior patterns and situations for surveillance analysis through activity recognition and  abnormal behavior detection.
Certain methods focus on classifying the most common, simple behavior patterns (linear, radial, etc.) in a given scene. However, most of these methods are designed for offline applications and tend to use a large number of training videos for offline learning of patterns for detecting common crowd behavior patterns~\cite{Shah_crowdPatterns}, normal and abnormal interactions~\cite{Mahadevan.anomaly.2010}, human group activities~\cite{Ni2009}. Other methods are designed for crowd analysis using a large number of web videos~\cite{mikel_dataDriven}. However, these techniques employ either manual selection methods or offline learning techniques for behavior analysis and therefore, cannot be used for interactive applications.
 All of these methods perform offline computations, and it is not clear whether they can be directly used for interactive applications.

\section{Public Policy Issues}
 As the legal, social and technological issues surrounding video surveillance are complex, this section provides a brief background of privacy issues especially in the context of our work.
 
 The ability of new camera and network technologies to identify, track, and investigate the activities of formerly anonymous individuals fundamentally changes the nature of video surveillance. While the various technological developments overlap, for these guidelines we conceive of four distinct types of surveillance technologies, each of which calls for differing rules and restrictions (a) observation technologies; (b) recording technologies; (c) tracking technologies; and (d) identification technologies. Even though our approach uses many of these technologies at the simultaneously, our technology may be employed to also mitigate the impact of surveillance on constitutional rights and values. Most of the debate around this is because most of the surveillance methods record facial features and details which can be tied to one's personal details, whereas, our research only looks at and learns for trajectories. No personal information is captured. In fact, the learned trajectories are also not stored in any database if the pedestrian is deemed ``harmless''.

We understand that public video surveillance systems have the potential to be used in ways that infringe on privacy and anonymity rights. Commentators often assume that there is ``no reasonable expectation of privacy'' in streets or parks or other areas open to view. As mentioned earlier, our approach takes a very safe and cautious approach at protecting personal information. We do not look at any visual cues relating to the person being tracked. Only pedestrian trajectory-level information is processed. 

It is often said that the risk of harm to constitutional rights and values posed by a public video surveillance system increases with its duration. The longer a system operates, the more activities and information it captures—permitting more and greater violations of privacy and anonymity and the correspondingly higher probability of public outcry and legal liability. Our approach is risk-free is this respect since, even though we use past trajectory to learn pedestrian behavior, the history we learn is only for a few seconds, and once the model is trained, that history is not used and deleted forever. As mentioned earlier, since we only use minimal information about the pedestrian (the trajectories), there is no threat to privacy and anonymity. There has been extensive research which models human gaits as potential identifiers of a person~\cite{chen2014average}, but so far no research has been able to establish a personal identification metrics with only trajectories.

\section{Trajectory Behavior Learning}

In this section, we present our interactive trajectory-level behavior computation
algorithm. 

\subsection{Terminology and Notation}
We first introduce the notation used in the remainder of the paper.

\textbf{Pedestrians:}
We use the term {\em pedestrian} to refer to independent individuals or human-like agents in the crowd. Their trajectories and movements are extracted by our algorithm using a realtime multi-person tracker. 

\textbf{State representation:} A key aspect of our approach is to compute the state of each pedestrian and each pedestrian cluster in the crowd video. Intuitively, the state corresponds to the low-level motion features that are used to compute the trajectory-level behavior features. 
In the remainder of the paper, we assume that all the agents are moving on a 2D plane. Realtime tracking of pedestrians is performed in the 2D image space and provides an approximate position, i.e. $(x,y)$ coordinates, of each pedestrian for each  frame of the video. In addition, we infer the velocities and intermediate goal positions of each pedestrian from the sequence of its prior trajectory locations. We encode this information regarding a pedestrian's movement at a time instance using a state vector. In particular, we use the vector $\mathbf{x}=[\mathbf{p} \; \mathbf{v} \; \mathbf{g}]^\mathbf{T}$, $\mathbf{x}\in\mathbb{R}^6$ to refer to a pedestrian's state. The state vector consists of three 2-dimensional vectors: $\mathbf{p}$ is the pedestrian's position, $\mathbf{v}$  is its current velocity, and $\mathbf{g}$ is the intermediate goal position. The intermediate goal position is used to compute the optimal velocity that the pedestrian would have taken had there been no other pedestrians or obstacles in the scene. As a result, the goal position provides information about the pedestrian's immediate intent. In practice, this locally optimal velocity tends to be different from $\mathbf{v}$ for a given pedestrian.
The state of the entire crowd, which consists of individual pedestrians, is the union of the set of each pedestrian's state $\mathbf{X}=\bigcup _i\mathbf{x_i}$.

\textbf{Pedestrian behavior feature:} The pedestrians in a crowd are typically in motion, and their individual trajectories change as a function of time. The behavior of the crowd can be defined using macroscopic or global flow features, or based on the gestures and actions of different pedestrians in the crowd.  In this paper, we restrict ourselves to trajectory-level behaviors or movement features per agent and per cluster, including current position, average velocity (including speed and direction), cluster flow, and the intermediate goal position. These features change dynamically. Our goal is to interactively compute these features from tracked trajectories, and then use them for behavior analysis.

\subsection{Overview}
\begin{figure*}[t]
  \centering
  \includegraphics[width = 1.0\linewidth]{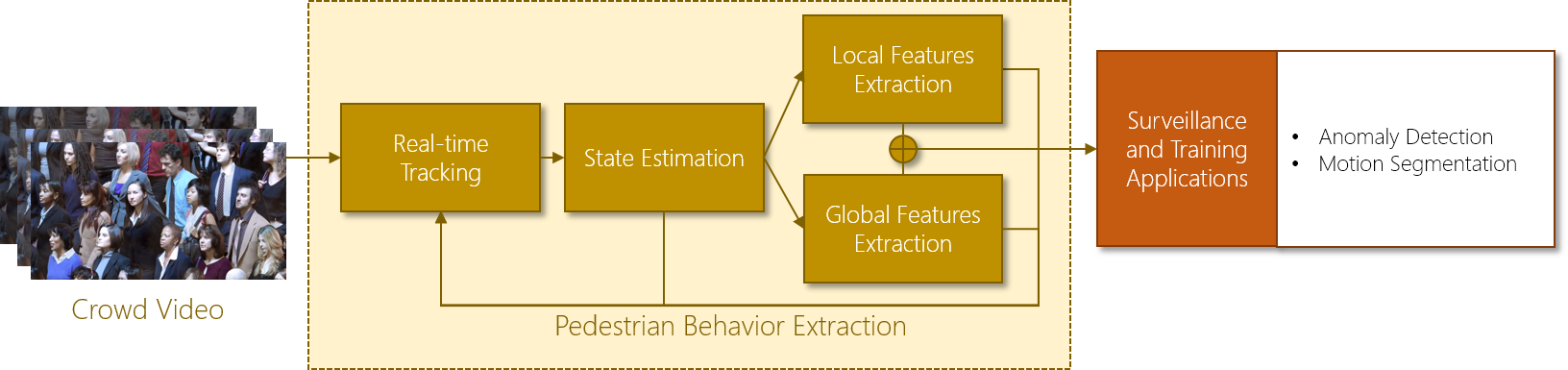}
  \caption{\textbf{Overview of our approach.} We highlight the different stages of our interactive algorithm: tracking, pedestrian state estimation, and behavior learning. The local and global features refer to individual vs. overall crowd motion features. These computations are performed at realtime rates for each input frame. }  \label{fig:overview}
\end{figure*}

Our overall approach consists of multiple components: a real-time multi-person tracker, state estimation, and behavior feature learning. 
One of our approach's benefits and its difference from prior approaches is that our approach does not require offline training using large number of training examples. As a result, it can be directly applied to any new or distinct crowd video. We extend our behavior learning and pedestrian tracking pipeline from~\cite{bera2014,kim2016interactive}.
Fig.~\ref{fig:overview} highlights these components. The input into our algorithm is one frame of real-world crowd video at a time, and our goal is to compute these behavior features for each agent from these frames. An adaptive multi-person or pedestrian tracker is used to compute the observed position of each pedestrian on a 2D plane, denoted as ($\mathbf{z}_0 \cdots \mathbf{z}_t$). Furthermore, we use new  state estimation and behavior-learning algorithms that can also compensate for the tracking noise  and perform robust behavior analysis.

We do not make any assumptions about the dynamics or the actual velocity of each agent in the crowd. 
Since we do not know the dynamics or true state of each agent, we estimate its state $\mathbf{x}$ from the recent observations for each pedestrian. We use a Bayesian inference technique to estimate the most likely state of each pedestrian in an online manner and thereby compute the state of the overall crowd,  $\mathbf X$.
Based on estimated real crowd states, we compute the trajectory behavior feature of each agent. These features are grouped together to analyze the behavior or movement patterns, and are also used for various training and surveillance applications.

 \noindent {\bf Interactive State Computation:} We use an online approach that is based on the current and recent states of each pedestrian. In other words, it does not require future knowledge or future state information for any agent. Because we estimate the state during each frame, our formulation can capture the local and global behavior or the intent of each agent.


\begin{figure}[ht]
  \centering
  \includegraphics[width = 1.0\linewidth]{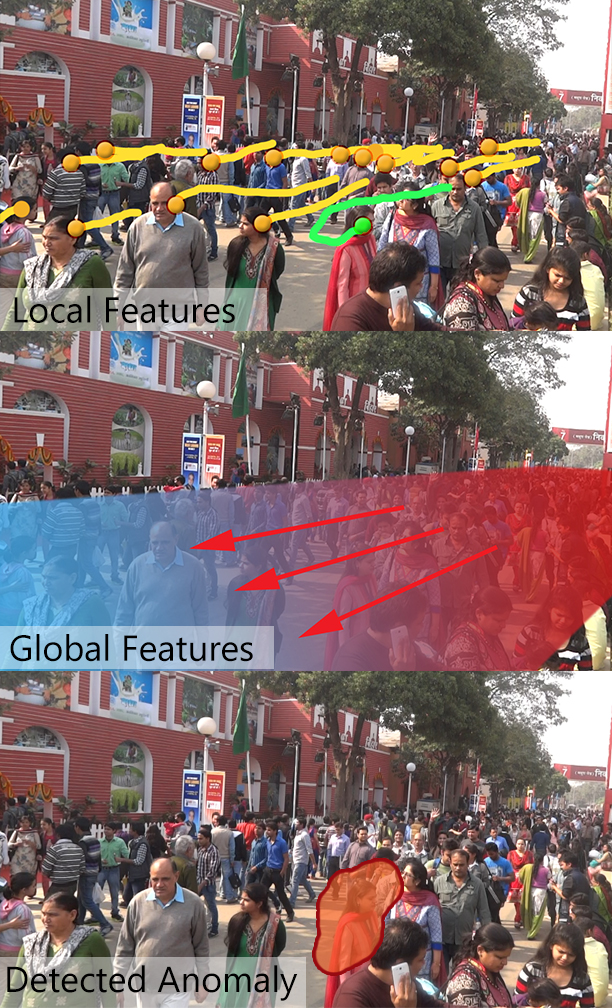}
  \caption{The \textbf{Anomaly Detection} In this example, we see that one pedestrian (marked with green) suddenly makes a U-turn (local feature) in a crowd where everyone is walking in a specific direction/field (global feature). Our system detects this as an anomaly. } \label{fig:AnomalyExample}
\end{figure}

Our approach uses a realtime multi-person tracking algorithm to extract the pedestrian trajectories from the video. There is considerable research in   computer vision literature on online or realtime tracking. 

To reliably estimate the motion trajectory in a dense crowd setting, we use RVO (reciprocal velocity obstacle)~\cite{van2011reciprocal} -- a local collision-avoidance and navigation algorithm -- as the non-linear motion model. For more details we direct our readers to ~\cite{bera2014,bera2015efficient,bera2015reach}

\begin{table}[h]
\centering
\resizebox{0.9\linewidth}{!}{%
\begin{tabular}{|l|l|l|l|}
\hline
\textbf{Video}   & \textbf{Density} & \textbf{Total Frames} & \textbf{BLT} \\ \hline
ARENA (01\_01)   & Low              & 1060                  & 0.002                                 \\ \hline
ARENA (01\_02)   & Low              & 890                   & 0.004                                 \\ \hline
ARENA (03\_05)   & Low              & 1440                  & 0.002                                 \\ \hline
ARENA (03\_06)   & Low              & 1174                  & 0.002                                 \\ \hline
ARENA (06\_01)   & Low              & 2941                  & 0.001                                 \\ \hline
ARENA (06\_04)   & Low              & 1582                  & 0.002                                 \\ \hline
ARENA (08\_02)   & Low              & 792                   & 0.002                                 \\ \hline
ARENA (08\_03)   & Low              & 746                   & 0.006                                 \\ \hline
ARENA (10\_03)   & Low              & 1173                  & 0.002                                 \\ \hline
ARENA (10\_04)   & Low              & 1188                  & 0.005                                 \\ \hline
ARENA (10\_05)   & Low              & 894                   & 0.004                                 \\ \hline
ARENA (11\_03)   & Low              & 329                   & 0.001                                 \\ \hline
ARENA (11\_04)   & Low              & 729                   & 0.002                                 \\ \hline
ARENA (11\_05)   & Low              & 666                   & 0.002                                 \\ \hline
ARENA (14\_01)   & Low              & 1081                  & 0.001                                 \\ \hline
ARENA (14\_03)   & Low              & 1242                  & 0.004                                 \\ \hline
ARENA (14\_05)   & Low              & 1509                  & 0.001                                 \\ \hline
ARENA (14\_06)   & Low              & 857                   & 0.002                                 \\ \hline
ARENA (14\_07)   & Low              & 1312                  & 0.004                                 \\ \hline
ARENA (15\_02)   & Low              & 917                   & 0.004                               
\end{tabular}}

\caption{Performance of trajectory level behavior learning on a single core for different benchmarks: We highlight the number of frames of extracted trajectories, the time spent in learning pedestrian behaviors (BLT - Behavior Learning Time (in sec)). Our learning and trajectory computation algorithms demonstrate interactive performance on these complex crowd scene analysis scenarios.}
\label{tab:dataset}

\end{table}

\subsection{Anomaly detection}
\label{sec:Anomaly}
Anomaly detection is an important problem that has been the focus of research in diverse research areas and applications. It  corresponds to the identification of pedestrians, events, or observations that do not conform to an expected pattern or to other pedestrians in a crowd dataset. Typically, the detection of anomalous items or agents can lead to improved automatic surveillance.
Anomaly detection can be categorized into two classes based on the scale of the behavior that is being extracted~\cite{kratz2009anomaly}: global anomaly detection and local anomaly detection. A global anomaly typically affects a large portion of, if not the entire, crowd and local anomaly is limited to an individual scale (for example, individuals moving against the crowd flow). We primarily use our trajectory-based behavior characteristics for local anomaly detection. In other words, we detect a few behaviors that are rare and are only observed in the video during certain periods. These periods can be as long as the length of the video or as short as a few hundred frames. In other words, we classify an anomaly as temporally uncommon behavior. For example, a person's behavior going against the flow of crowds may be detected as an anomaly at one point, but the same motion may not be detected as an anomaly later in the frame if many other pedestrians are moving in the same direction. 

For anomaly detection we compare the distance between the local and global pedestrian features of every pedestrian (we refer the readers to read~\cite{Bera_2016_CVPR_Workshops} for more technical details and analysis. ). When an anomaly appears in a scene, the anomaly features typically tend to be isolated in the cluster of which it is a part. In other words, the pedestrian's motion will be different from that of the surrounding crowd. If the Euclidean distance between the \textit{global} and \textit{local} feature is more than a threshold value, we classify it as an anomaly.

\begin{eqnarray}
dist(\mathbf{b}^l, \mathbf{b}^g) > Threshold
\end{eqnarray}

This threshold is a user-tunable parameter. If this threshold is set low, the sensitivity of the anomaly detection will increase and vice-versa.

\section{Quantitative Results}

We compare the accuracy of our motion segmentation and anomaly detection methods using the quantitative metrics presented in Table 1 and Table V, as described in Li et al. ~\cite{LiCrowdedSceneAnalysis2015}. Table 1 in ~\cite{LiCrowdedSceneAnalysis2015} provides a true detection rate for motion pattern segmentation. It is based on the criterion that the approach successfully detected the regions containing the moving pedestrians. Although we cannot directly compare the numbers with pixel-based performance measures, MOTP values (Table 1) can be an indirect measure for the true detection rate motion segmentation. Compared to the values range of \textbf{0.4-1.0} in [15], the corresponding values computed by our approach are in the range of \textbf{0.7-0.8} in terms of detecting moving pedestrians, even for unstructured videos. These numbers indicate that the performance of our method is comparable to the state of the art.

\begin{figure}[htb]
\centering
\subfloat[]{\includegraphics[width=0.43\linewidth]{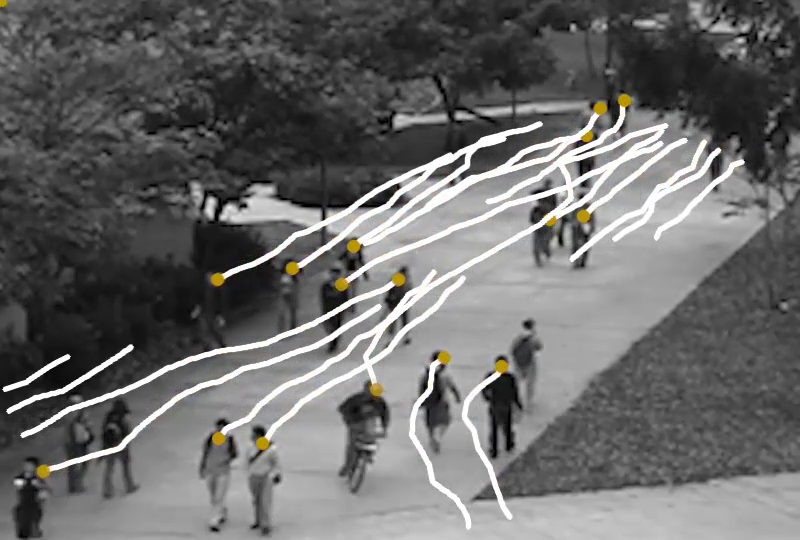}}
\subfloat[]{\includegraphics[width=0.48\linewidth]{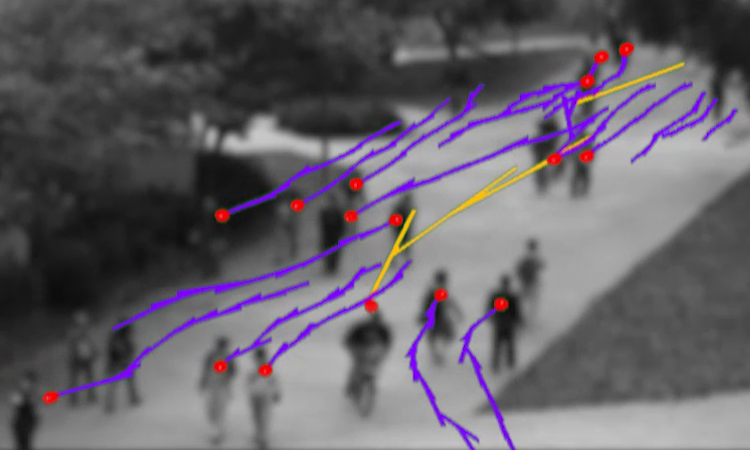}}
\caption{\textbf{Anomaly Detection}. We also evaluate other datasets like UCSD. Trajectories of 63 real pedestrians are extracted from a video. One person in the middle walks against the flow of crowd. Our method can capture the anomaly of this pedestrian's behavior or movement by comparing the behavior features with those of other pedestrians. }
\label{fig:Anomaly}
\end{figure}

Fig.~\ref{fig:Anomaly} shows the results of anomaly detection in different crowd videos. 
\textbf{879-38 video dataset~\cite{mikel_dataDriven}}: The trajectories of $63$ pedestrians are extracted from the video. One person in the middle is walking against the flow of pedestrians through a dense crowd. Our method can distinguish the unique behavior of this pedestrian by comparing its behavior features with those found by methods. In \textbf{UCSD-Peds1-Biker} and \textbf{UCSD-Peds1-Cart} benchmarks, our method is able to distinguish parts of the trajectories of the biker and the cart because their speeds were noticeably different from those of other pedestrians.

Apart from ARENA, we evaluated the accuracy of the anomaly detection algorithm on the UCSD PEDS1 dataset  ~\cite{Mahadevan.anomaly.2010} and compared it with Table V in Li et al. ~\cite{LiCrowdedSceneAnalysis2015} in Table 2. 

Our method successfully detected the following anomalies in the ARENA - \textit{Person checking vehicle}, \textit{different motion pattern}, \textit{person on a bike}, \textit{push and run}, \textit{abnormal motion near vehicle}, \textit{man touching vehicle}, \textit{hit and run}, \textit{suddenly people running} and \textit{possible mugging}.

\begin{table}
\centering
\label{my-label}
\resizebox{\linewidth}{!}{%
\begin{tabular}{|c|c|c|c|c|c|c|}
\hline
\multirow{2}{*}{{\bf Reference}} & \multirow{2}{*}{{\bf Dataset}} & \multicolumn{5}{c|}{{\bf Performance}}                                                   \\ \cline{3-7} 
                                 &                                & {\bf \it Area under ROC Curve} & {\bf \it Accuracy} & {\bf \it DR} & {\bf \it Equal Error Rate} & {\bf \it Online/Offline} \\ \hline
{\bf Our Method}                 & \multirow{5}{*}{UCSD}          & {\bf 0.873}    & {\bf 85\%}    & {\bf -}      & {\bf 20\%}    & {\bf Online}             \\ \cline{1-1} \cline{3-7} 
Wang 2012                        &                                & 0.9            & -             & 85\%         & -             & Offline                  \\ \cline{1-1} \cline{3-7} 
Cong 2013                        &                                & 0.86           & -             & -            & 23.9          & Offline                  \\ \cline{1-1} \cline{3-7} 
Cong 2012                        &                                & 0.98-0.47      & 46\%          & 46\%         & 20\%          & Offline                  \\ \cline{1-1} \cline{3-7} 
Thida 2013                       &                                & 0.977          & -             &              & 17.8\%        & Offline                  \\ \hline
Our Method                       & 879-44                         & {\bf 0.97}     & {\bf 80\%}    & {\bf -}      & {\bf 13\%}    & {\bf Online}             \\ \hline
Our Method                       & ARENA                         & {\bf 0.91}     & {\bf 76\%}    & {\bf -}      & -    & {\bf Online}             \\ \hline
\end{tabular}}
\caption{Comparison of Anomaly Detection techniques. Our method has comparable results with the state of the art offline methods in anomaly detection.}
\end{table}

\begin{table}[]
\centering
\begin{tabular}{|l|l|l|}
\hline
\textbf{Video Name} & \textbf{Camera ID} & \textbf{Threat Level} \\ \hline
11\_03              & TRK\_RGB\_1        & High                  \\ \hline
15\_02              & TRK\_RGB\_1        & High                  \\ \hline
22\_02              & ENV\_RGB\_3        & High                  \\ \hline
14\_06              & TRK\_RGB\_1        & Medium                \\ \hline
15\_06              & TRK\_RGB\_1        & Medium                \\ \hline
14\_07              & TRK\_RGB\_1        & Low                   \\ \hline
10\_04              & TRK\_RGB\_1        & Low                   \\ \hline
06\_01              & TRK\_RGB\_1        & Low                   \\ \hline
10\_05              & TRK\_RGB\_1        & Low                   \\ \hline
\end{tabular}
\caption{Details of the anomalies detected in the ARENA Dataset.}
\end{table}

{\small
\bibliographystyle{aaai}
\bibliography{refs}
}
\end{document}